\documentclass{article}
\usepackage{ijcai22}

\usepackage{ragged2e}
\usepackage{booktabs}
\usepackage{bm}
\usepackage{xcolor}
\usepackage{arydshln}
\usepackage{CJKutf8}
\usepackage{enumitem}

\usepackage{times}
\usepackage{soul}
\usepackage{url}
\usepackage[hidelinks]{hyperref}
\usepackage[utf8]{inputenc}
\usepackage[small]{caption}
\usepackage{graphicx}
\usepackage{amsmath}
\usepackage{amsthm}
\usepackage{booktabs}
\usepackage{algorithm}
\usepackage{algorithmic}
\usepackage{authblk}

\title{You talk what you read: Understanding News Comment Behavior \\ by Dispositional and Situational Attribution}

\author[$\dagger$]{\textbf{Yuhang Wang}}
\author[$\dagger$]{\textbf{Yuxiang Zhang}}
\author[$\ddagger$]{\textbf{Dongyuan Lu}}
\author[$\dagger$\thanks{Corresponding author}]{\textbf{Jitao Sang}}
\affil[$\dagger$]{Beijing Key Lab of Traffic Data Analysis and Mining \authorcr
          Beijing Jiaotong University, Beijing, China \authorcr
          \{yhangwang, yuxiangzhang, jtsang\}@bjtu.edu.cn}
\affil[$\ddagger$]{School of Information Technology and Management \authorcr
          University of International Business and Economics, Beijing, China\authorcr
          ludy@uibe.edu.cn}

\begin{document}

\maketitle

\begin{abstract}
Many news comment mining studies are based on the assumption 
that comment is explicitly linked to the corresponding news. 
In this paper, we observed that users' comments are also heavily influenced by
their individual characteristics embodied by the interaction history. 
Therefore, we position to understand news comment behavior by considering 
both the dispositional factors from news interaction history, 
and the situational factors from corresponding news. 
A three-part encoder-decoder framework is proposed to model the generative process of news comment. 
The resultant dispositional and situational attribution contributes to understanding user focus and opinions, 
which are validated in applications of reader-aware news summarization and news aspect-opinion forecasting.
\end{abstract}

\section{Introduction}
Increasingly more people express their opinions on news articles through online services recently, 
such as news portals and microblogs.
The resulting vast number of comments clearly reflect thoughts and feelings of 
individuals. 
Mining these comments thus has important applications
with practical socio-political and economical benefits.~\cite{hou2017learning}
~\cite{boltuvzic2014back}

Existing related work has explored comment data in different scenarios.  
One typical line of studies ~\cite{pontiki2016semeval}~\cite{Peng2020KnowingWH}~\cite{yan2021unified} concentrated on Sentiment Analysis tasks, 
especially Aspect-based Sentiment Analysis(ABSA)
which aims to identify the aspect term, its corresponding sentiment polarity, and the opinion term.
Another research line was based on the interaction between news and comments.
A fundamental assumption for these studies is that 
comments have clear correspondence with certain aspects of the corresponding news. 
Based on the explored news-comments correspondence, 
~\cite{hou2017learning} aligned comments to news topics, which improves readers' news browsing experience,
~\cite{yang2020predicting} introduced a new task which leverages reading and commenting history 
to predict a user's future opinions to unseen news,
and researchers also developed automatic news commenting algorithms to encourage user engagement and interactions~\cite{qin2018automatic}.
For example, ~\cite{wang2021generating} incorporated reader-aware factors to generat diversified comments.
~\cite{li2019coherent} modeled the news as a topic interaction graph to capture the main point of the article, 
which enhances the correspondence between generated comments and news.

However, sometimes comments not explicitly link to the corresponding news. 
Table \ref{tab:news-comment-example}(top)
illustrates an example of news and associated comments that 
\emph{Wuhan's patients with COVID-19 in the hospital are fully recovered.}
\begin{table}[t]
    \setlength{\abovecaptionskip}{0.1cm}
    \centering
    \begin{tabular}{p{0.95\columnwidth}}
    \toprule[1.5pt]
    \textbf{Title:} Live screen! The last batch of COVID-19 patients in Wuhan have been discharged from hospital.\\
    \textbf{Body:} ``Finally!" On \textcolor{cyan}{\textbf{April 26}}, a COVID-19 patient surnamed Ding was discharged from Wuhan Pulmonary 
    Hospital, and \#all hospitalized COVID-19 patients in Wuhan were cleared\#. 
    Netizen: The day of wuhan's recovery is the \textcolor{purple}{\textbf{Chinese New Year}}.\\
    \midrule[0.3pt]
    \textbf{Comment:}\\
    \textbf{User-1:} Congratulations! This is a \textcolor{cyan}{\textbf{memorable day}}!\\
    \textbf{User-2:} The last batch of \textcolor{orange}{\textbf{Jiangsu}} Medical teams to aid Hubei went home. Thank you.\\
    \textbf{User-3:} Great \textcolor{purple}{\textbf{China}}!\\
    \midrule[1.2pt]
    \textbf{Partial news-comment history of User-2:}\\
        \textbf{news:}
        \textcolor{orange}{\textbf{Jiangsu}} launched a level 1 public health emergency response to prevent the spread of the virus.\\
        \textbf{comment:}
        Each student has been screened in my daughter's school today.\\
        \hdashline[2.5pt/5pt]
        \hdashline[2.5pt/5pt]
        \textbf{news:}
        The second group of medical workers from \textcolor{orange}{\textbf{Jiangsu}} province relay to Wuhan.\\
        \textbf{comment:}
        Salute to the most beautiful people.\\
        \hdashline[2.5pt/5pt]
        \hdashline[2.5pt/5pt]
        \textbf{news:}
        Four cases of pneumonia caused by the COVID-19 were confirmed in \textcolor{orange}{\textbf{Jiangsu}} province, 
        all with recent travel history to Wuhan.\\
        \textbf{comment:}
        Suzhou is the first city in \textcolor{orange}{\textbf{Jiangsu}} province to find (confirmed cases).\\
    \bottomrule[1.5pt]
    \end{tabular}
    \caption{
        Comment from NetEase: (top) an example news with comments; 
        (bottom) user2's reading and commenting history. 
        News and comments are originally in Chinese and translated to English.
    }
    \label{tab:news-comment-example}
\end{table}
The comments about ``memorable day'' and ``China'' link entities to the corresponding news, 
however, the comments about ``Jiangsu'' from user-2 has no explicit correspondence.
By retrieving user-2's news interaction (i.e., reading and commenting) history as shown in Table 1 (bottom), 
we find that he/she heavily concerns about the topics related to ``Jiangsu'', 
which gives rise to the above comment combining the news' topics on COVID-19 and user-2's individual focus on ``Jiangsu''. 

To further investigate whether the above phenomenon is common,
we study how comment entities distribute in corresponding news 
and their interaction history on 85,179 comments from 1,275 users on NetEase News\footnote{\small{https://news.163.com/}}.
Table \ref{tab:DIST_stst} shows the distribution of comments with respect to 
the entity linking. 
We observe that 34\% comments have no key entities clearly linked to the corresponding news,
among which 
nearly 2/3 (20\%) 
appear only in users' interaction history.
The result demonstrates that user comment is related to not only the corresponding news 
but also user's individual characteristics embodied by the interaction history. 

Inspired by this, in this paper, 
we position to understand news comment behavior by modeling both user's interaction history and the corresponding news. 
According to the attribution theory ~\cite{heider2013psychology}~\cite{Heider1944AnES},
human behavior attribution can be divided into dispositional attribution(e.g. emotions, attitudes, abilities, etc.)  
and situational attribution(e.g. event or external pressure etc.). 
Intuitively, in the news comments scenario, 
mining interaction history and the corresponding news contribute to dispositional and situational attribution respectively.
Based on the above analysis and conclusion, we develop a three-part generative framework named DS-Attributor
to understand news comment behavior by Dispositional and Situational Attribution.
The first part is Dispositional Factor Encoder to model individual characteristics
with both aspect and opinion user topic preferences.
The second part is Situational Factor Encoder 
exploiting the user-derived aspect topics from dispositional factor to detect the focused aspects of specific news.
Finally, the mined opinion topics of dispositional factor and the detected situational factor
are integrated into the Dynamic Comment Decoder module to generate comments.
\begin{table}[t]
    \setlength{\abovecaptionskip}{0.1cm}
    \centering
    \begin{tabular}{lr}
    \toprule
     Entities of comments appear in & Percentage \\
    \midrule
     only corresponding news             & 21\% \\
     corresponding news \& history  & 55\% \\
     only history      & 20\% \\
     neither & 14\% \\ 
    \bottomrule
    \end{tabular}
    \caption{Distribution of news-comment correspondence}
    \label{tab:DIST_stst} 
\end{table}

    
  
\textbf{Contributions.} We summarize the main contributions of this paper as follows:
\begin{itemize}
    \item We position the problem of understanding
     news comment behavior by both situational and dispositional attribution.
    \item We propose a novel encoder-decoder framework to model the comment generation process 
    by combining the comment history and corresponding news. 
    \item The resultant dispositional and situational attribution 
    is validated to enable applications like news aspect-opinion forecasting and reader-aware news summarization.
\end{itemize}

\section{Notations and Problem Definition}
Our goal is to understand news comment behavior by dispositional and situational attribution through a generative framework.
Specifically, given a user, the model needs to use his/her historical comments to mine dispositional factors and
detect the situational factors from a specific piece of news, 
then generates comment with the mined dispositional and situational factors.  
Let $\mathcal{U}=\{u_{1}, u_{2},...,u_{N}\}$ denote a set of users. 
For each user $u_{n}\in\mathcal{U}$,
assume $\mathcal{P}_{u_n}=\{Y_{u_n,1}, Y_{u_n,2},\cdots,Y_{u_n,t},\cdots,Y_{u_n,T_{u_n}}\}$ 
to include all comments $u_n$ posted before timestep $T_{u_n}$.
The comment $Y_{u_n,t}$ denotes a sequence of words as $\{y_1,y_2,\cdots,y_l\}$,
where $y_{i}\in\mathcal{V}$, and $l$ is the number of words in $Y_{u_n,t}$.
Let $X$ denotes a piece of news which haven't been read by $u_{n}$ before $T_{u_n}$ as
$\{t,b_1,b_2,\cdots,b_m\}$ includes news title $t$ and $m$ sentences of news body.
Based on the above notations, we formally define the problem as follows:

\noindent \textbf{Problem 1(Dispositonal and Situational Comment Attribution)}
Given the historical comments $\mathcal{P}_{u_n}$ of user $u_{n}\in\mathcal{U}$ 
and a specific piece of news $X$, the goal of Dispositonal and Situational Comment Attribution is:
(1) to mine dispositional factor which includes preferences regarding both aspect and opinion topics from $\mathcal{P}_{u_n}$,
(2) to detect situational factor from news $X$, 
(3) to generate comment $Y$ on news $X$ based on dispositional and situational factors.  

\begin{figure*}[htbp]
    \centering 
    \setlength{\abovecaptionskip}{0.1cm}
    \includegraphics[scale=0.57]{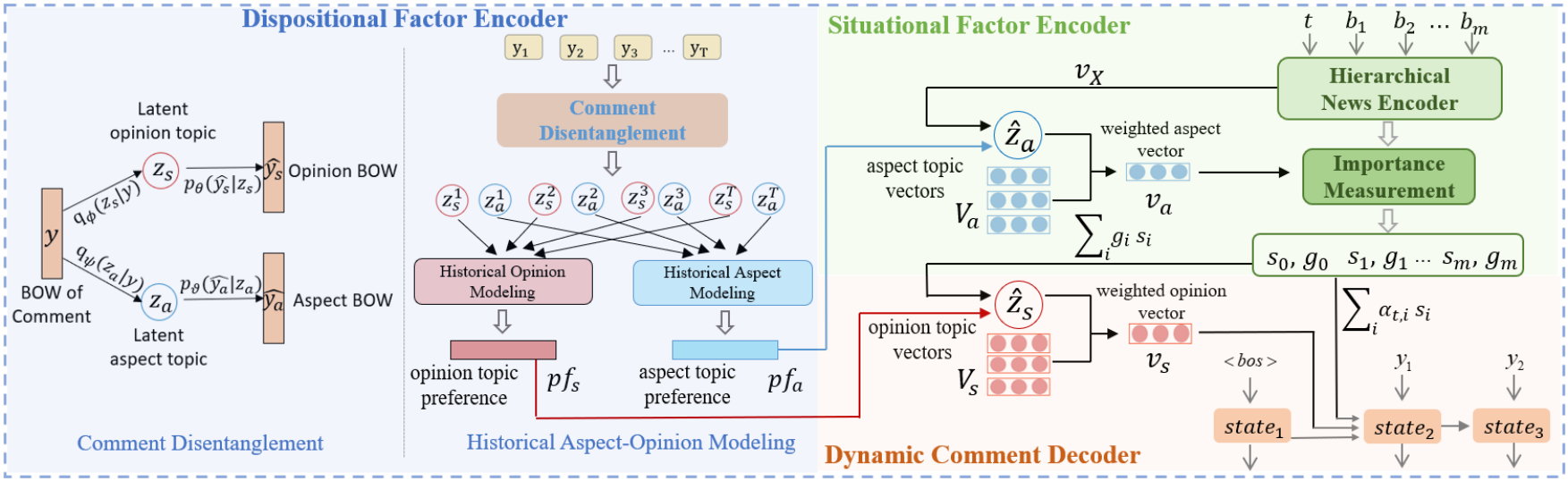}
    \caption{Overall architechture of the proposed DS-Attributor.}
    \label{fig:framework}
    \vspace{-2mm}
\end{figure*} 
\section{Methodology}
We present the overall framework DS-Attributor in Figure \ref{fig:framework}, which includes three main modules.
The Dispositional Factor Mining Encoder aims at modeling dispositional factors
involving with both aspect and opinion topic preferences $pf_a$, $pf_s$ from users' historical comments (see Section \ref{sec:Dispositional Factor Mining Module}). 
In Situational Factor Encoder, given aspect topic preference $pf_a$ and news $X$,
the goal is to get representation $s_i$ for each news sentence, 
and measure the corresponding importance $g_i$ by a weighted aspect vetor $v_a$ (see Section \ref{sec:Situational Factor Encoder}).
Finally, in Dynamic Comment Decoder, users opinion vector $v_s$ is obtained  
and incorporated with $s_i$ and $g_i$ of each sentence to generate the observed comment (see Section \ref{sec:Dynamic Decoder}).
We will elaborate the details of each module below.

\subsection{Dispositional Factor Encoder}
In this subsection, our goal is to mine dispositional factor from users' historical comments.
The comment is mainly composed of aspect and opinon terms~\cite{pontiki2016semeval},
for example, in the sentence \emph{Great China!}, the aspect term is ``China'', 
and the opinion term is ``Great''.
Therefore, we model the dispositional factor as the user's preferences of aspect and opinion.

\noindent \textbf{Comment Disentanglement.}
We first pretrain a Comment Disentanglement module based on Neural Topic Model ~\cite{dieng2020topic},
which can be used to extract aspect and opinion topic distributions from the comment.
In addition, the aspect topic vectors $V_a \in \mathcal{R}^{k_a \times d}$ 
and opinion topic vectors $V_s \in \mathcal{R}^{k_s \times d}$ also can be obtained,
where $d$ is the dimension of topic vetors, $k_a$ and $k_s$ are numbers of aspect and opinion topic respectively. 
Specifically, as shown in the left of Figure \ref{fig:framework},
for each comment, we employ Bag-of-Words(BOW) feature vector $y$ to represent it.
Then we use two parallel VAE-based structures to reconstruct aspect BOW target $\widehat{y}_a$ and
opinion BOW target $\widehat{y}_s$ respectively.
$\widehat{y_{a}}$ and $\widehat{y_{s}}$ are defined as:
\begin{align}
    \widehat{y}_{a,i} &= \left\{
        \begin{array}{rl}
            y_{i} & \text{if the word corresponding to $y_{i}$}\\
                  & \text{is an entity},\\
            0 &  \text{else }.\\
        \end{array}
    \right.,                       \\
    \widehat{y}_{s,i} &= \left\{
        \begin{array}{rl}
            y_{i} & \text{if the word corresponding to $y_{i}$} \\
                  & \text{is an adjective or emotional word},\\
            0 &  \text{else }.\\
        \end{array}
    \right.,
\end{align} 
where $y_i$, $\widehat{y}_{a,i}$, $\widehat{y}_{s,i}$ are elements of 
$y$, $\widehat{y}_{a}$, $\widehat{y}_{s}$ respectively.
During inference, 
given a comment BOW feature vector,
we can get both aspect topic distribution $z_{a}$ 
and opinion topic distribution $z_{s}$.

\noindent \textbf{Historical Aspect-Opinion Modeling.}
For each user $u_n$, the historical sequences of both aspect and opinion topic distributions
can be obtained by performing comment disentanglement on the historical comments $\mathcal{P}_{u_n}$. 
Denote $[z^1_a,z^2_a,\cdots,z^t_a,...,z^T_a]$ , $[z^1_s,z^2_s,\cdots,z^t_s,...,z^T_s]$
as the derived historical sequences of aspect and opinion topic distributions.
We introduce two different LSTMs to process the above distribution sequences 
and output user preferences of aspect topic $pf_a$ and opinion topic $pf_s$ respectively.
Specifically, the $t$-th hidden states are given by:
\begin{align}
    pf^t_a &= LSTM_{a}(z^t_a, pf^{t-1}_a), \\ 
    pf^t_s &= LSTM_{s}(z^t_s, pf^{t-1}_s).
\end{align}
After recursive updating, 
we encode user's propositional preference in user-aspect topic $pf_a$ and user-opinion topic $pf_s$.
\label{sec:Dispositional Factor Mining Module}
\subsection{Situational Factor Encoder}
\label{sec:Situational Factor Encoder}
In this subsection, our goal is to detect situational factor from the news.
Since not all sentences contribute equally to motivate users to comment,
we introduced an attention-based method using weighted aspect vector $v_a$
to measure importance of news sentences with respect to a specific user.

\noindent \textbf{Hierarchical News Encoder.}
Firstly, the news $X$ is embedded as $v_X$ by a hierachical news encoder. 
Assume the news $X$ contains $L$ sentences and each sentence contains $N$ words.
$w_{i,t}$ with $t \in [1,N]$ represents the words in the $i$th sentence.
Given a sentence, we first use Bi-LSTM to get word embeddings from both directions for words as:
\begin{equation}
    \begin{aligned}
        s_{i,t} &= \text{Bi-LSTM}(w_{i,t},s_{i,t-1}). \\
    \end{aligned}
\end{equation}
An attention mechanism ~\cite{vaswani2017attention} is introduced to aggregate the representation of those informative words 
to form a sentence vector as follows:
\begin{align}
    s_i =& \sum\nolimits_{t}\alpha_{i,t}s_{i,t}, \\
    \alpha_{i,t} =&  softmax(u^\top_{i,t}{u_w}), \\
    u_{i,t} =& tanh({W_w}s_{i,t} + {b_w}).
\end{align}
To get the news embedding, we aggregate the sentence vectors by attentive pooling as:
\begin{align}
    v_X =& \sum\nolimits_{i}\beta_is_i, \\
    \beta_{i} =&  softmax(u^\top_{i}{u_s}),\\
    u_i =& tanh({W_s}s_i + {b_s}).
\end{align}

\noindent \textbf{Importance Measurement.}
Since $pf_a$ reflects the user's aspect preference for news content,
we employ it to analyse the importance of sentences in the news.
Specifically, we first take $pf_a$ and $v_X$ as inputs to predict aspect topic distribution $\widehat{z}_a$,
and the weighted aspect vector $v_a$ is then calculated as:
\begin{align}
    \label{equ:aspect vector}v_a =& \sum_{j}\nolimits V_{a}(j)\widehat{z}_{a}(j), \\
    \label{equ:aspect distribution} \widehat{z}_a =& softmax({W_z}[pf_a;v_X]), 
\end{align}
where $V_{a}(j)$ is the $j$th aspect vector obtained from Comment Disentanglement module.
Note that in the training stage, the true aspect topic distribution $z_a$ is available.
During inference, we predict aspect topic distribution by Eqn.~\eqref{equ:aspect distribution}.
So in order to learn ${W_z}$ during the training stage, 
a KL term is added into the final loss function:
\begin{align}
    \bm{\mathcal{L}_a}=D_{KL}(\widehat{z}_a||z_a).
\end{align}
Similarly, \bm{$\mathcal{L}_s}=D_{KL}(\widehat{z}_s||z_s)$ is used to constrain $\widehat{z}_s$.
Then, we employ another attention mechanism to measure importance of sentences using aspect vector.
The importance score $g_i \in [0,1]$ for each sentence can be obtained by:
\begin{align}
    g_{i} =&  softmax(u'^\top_{i}{u'_s}),\\
    u'_{i} =& tanh({W_g}s_{i} + {b_g}).
\end{align}

\subsection{Dynamic Comment Decoder}
Considering the opinion topic preference,
we design a comment decoder to dynamically integrate
the opinion vector and the news context vector to generate comments.
The formula of the decoder is defined by:
\begin{align}
    state_t = LSTM_{dec}([c^X_t; e(y_t-1); M_t], state_{t-1}),
\end{align}
where $[\cdot;\cdot]$ denotes vector concatenation.
The comment word is then sampled from output distribution based on the concatenation of decoder
state as:
\begin{equation}
    \tilde{y}_t \sim softmax({W_y}(state_t)).
\end{equation}
During training, cross-entropy loss \bm{$\mathcal{L}_{ce}$} is employed as the optimization objective.
The decoder takes the embedding of the previously decoded word $e(y_t-1)$ , the context vector
$c^X_t$ and the dynamic opinion state $M_t$ as input to update its state.
The context vector $c^X_t$ is a weighted sum of encoder’s sentence representations calculated by:
\begin{align}
    c^X_t = & \sum\nolimits_{i}\alpha_{t,i}s_i, \\
    \alpha_{t,i} = & \frac{g_i \odot e_{t,i}}{\sum_{j}g_{j} \odot e_{t,j}}, \\
    e_{t,i} = & softmax(state_{t-1}{{W_\alpha}}s_i).  
\end{align} 
The dynamic opinion state $M_t$ is initialized by the opinion vetor $v_s$ which is calculated similar with $v_a$(see Eqn. \eqref{equ:aspect distribution}), 
and decays by a certain amount at each time step. 
This process is described as:
\begin{align}
    M_t = & g^u_t \odot M_{t-1}, \\
    g^u_t = & sigmoid({W_o}(state_t)), \\
    M_{0} = & v_s, \\
    \widehat{z}_s =& softmax({W_s}[pf_s;\sum\nolimits_{i}g_is_i]),
\end{align}
where $\odot$ denotes element-wise multiplication.
Once the decoding process is completed, the opinion state is expressed on context vector completely,
and the comment is generated.\par
\label{sec:Dynamic Decoder}
The above three modules are jointly trained with the following overall loss function
\begin{align}
    \mathcal{L} = \mathcal{L}_{ce} + \lambda_1\mathcal{L}_{a} + \lambda_2\mathcal{L}_{s},
    \label{equ:loss}
\end{align}
where $\lambda_1$, $\lambda_2$ are hyperparameters to balance between different modules.
Then we jointly train all components according to Eqn.~\eqref{equ:loss}.
After dispositional and situational comment attribution,
we can obtain $\widehat{z}_a$, $\widehat{z}_s$, $g_i$ 
and the decoder attention $e_i(\text{the mean of }e_{t,i})$ to support the following experiments and applications.


\section{Experiments}
\subsection{Experiments Setup} 
\noindent \textbf{Datasets.}
Since existing news datasets do not include user interaction history to satisfy situational attribution, 
we construct a new dataset DS-News from NetEase News, one of the most popular online news platfrom in China. 
We set 10 random users and crawl users who commented on the same news article using Breadth First Search.
For a specific user, we crawl his/her interaction history which consists of a sequence of news-comment pairs.
After removing users with too short interaction history, 1,275 examined users are collected with totally 124,918 comments.
Table~\ref{tab:news-comment-example}(top) visually shows a news comment instance.
The statistics of DS-News are summarized as shown in Table~\ref{tab:dataset-stat}. 
\begin{table}[t]
    \centering
    \setlength{\abovecaptionskip}{0.1cm}
    \begin{tabular}{rr}\\
    \toprule
    Dataset attributes & number\\
    \midrule
    Total number of users & 1,275\\
    Total number of news & 97,937\\
    Total number of comments & 124,918\\
    Avg. length of user histories & 97.63\\
    Avg. number of news words & 382.77\\
    Avg. number of comment words & 17.10\\
    \bottomrule
    \end{tabular}
    \caption{\label{tab:dataset-stat} Statistics of DS-News}
    \vspace{-3mm}
\end{table}

\noindent \textbf{Compared Methods.}
In order to evaluate the effectiveness of the proposed DS-Attributor, 
we implemented the following baselines and DS-Attributor variants for comparison 
in terms of the news comment generation task.
\begin{itemize}[leftmargin=*]
\setlength{\itemsep}{0pt}
\setlength{\parsep}{0pt}
\setlength{\parskip}{0pt}
\item \emph{Seq2seq}(\cite{qin2018automatic}): this model follows the framework of seq2seq model with attention.
We use the title together with the content as input. 
\item \emph{Hierarchical-Attention}(\cite{yang2016hierarchical}): this model takes all the content sentences as input and applies hierarchical 
attention as the encoder to get the sentence vectors and document vector. 
A RNN decoder with attention is applied. The document vector is used as the initial state for RNN decoder. 
\item \emph{Graph2seq}(\cite{li2019coherent}): this model constructs the input news as a topic interaction graph,
and it takes the GCN as encoder and the LSTM as decoder to generate news comment.
\item \emph{DS-Attributor(w/o IM):} DS-Attributor without Importance Measurement. 
\item \emph{DS-Attributor(w/o OV):} DS-Attributor without integrating opinon vector $v_s$.
\end{itemize}

\noindent \textbf{Evaluation protocols.}
We use BLEU-1, BLEU-2 ~\cite{papineni2002bleu}, ROUGE-L ~\cite{lin2004rouge}, 
CIDEr\cite{vedantam2015cider} and METEOR ~\cite{banerjee2005meteor} as metrics to evaluate
the performance of different models. 
A popular NLG evaluation tool nlg-eval \footnote{\small{https://github.com/Maluuba/nlg-eval}} is used to compute these metrics.

\noindent \textbf{Implementation details.} For pretraining Comment Disentanglement,
we use a vocabulary with the top 20k frequent words in the entire data. 
The number of aspect topics and opinion topics are set to 40 and 6 respectively. 
The dimensions of the latent topic vectors are both set to 300. 
We pretrain the model using Adam~\cite{kingma2014adam} with learning rate 0.001.
For Historical Aspect-Opinion Modeling, we use different two-layer LSTMs with hidden size 64 
to model aspect and opinion topics respectively.
For sentence encoder, we use a two-layer Bi-LSTM with hidden size 128.
For importance measurement, we employ attention mechanism with hidden size 256.
We use a two-layer LSTM with hidden size 512 as decoder.
For our method, 
$\lambda_1$ and $\lambda_2$ are both set to 0.4 
\footnote{More implementation details and results by tuning the hyperparameters are available in the supplementary material.}.
The batch size is set to 64.
Those parameters are optimized by Adam optimizer with learning rate 0.001 and trained for 200 epochs with learning rate decay.

\begin{table}[t]
    \centering
    \setlength{\abovecaptionskip}{0.1cm}
    \setlength\tabcolsep{1.8pt}
    \resizebox{0.48\textwidth}{!}{
    \begin{tabular}{rrrrrr}
      \toprule 
       Methods & BLEU-1 & BLEU-2 & ROUGE-L & METEOR & CIDEr\\ 
      \midrule
        Seq2seq & 0.101 & 0.021 & 0.091 & 0.046 & 0.029\\
        Graph2seq & 0.108 & 0.020 & 0.093 & 0.044 & 0.023\\
        Hierarchical-Attention & 0.102 & 0.022 & 0.092 & 0.044 & 0.037\\
        DS-Attributor(w/o IM) & 0.118 & 0.027 & 0.103 & 0.051 & 0.034\\
        DS-Attributor(w/o OV) & 0.121 & 0.027 & 0.094 & 0.053 & 0.034\\
        DS-Attributor & \textbf{0.125} & \textbf{0.029} & \textbf{0.108} & \textbf{0.054} & \textbf{0.039}\\
      \bottomrule
    \end{tabular}
    }
    \caption{Evaluation results in terms of news comment generation}
    \label{tab:result}
    \vspace{-2mm}
\end{table}

\subsection{Quantitative Experimental Results}
Quantitative evaluation results are shown in Table \ref{tab:result}.
The proposed DS-Attributor outperforms the baselines on all 5 evaluation metrics.
This demonstrates the advantage of exploring the dispositional factors in modeling news comment behaviors. 
In Table~\ref{tab:aspect topics} and Table~\ref{tab:opinion topics}, we illustrate some example aspect and opinion topics discovered from news interaction history. 
We can see that aspect topics describe different focuses and interests of users, 
and opinion topics help understand user sentimental preferences. 
Among the baseline methods, Hierarchical-Attention generally performs better than Seq2seq and Graph2seq.
A possible reason is that 
Hierarchical-Attention captured and aggregated the key information in the news through hierarchical attention mechanism. 

On all 5 evaluation metrics, 
DS-Attributor achieved superior performance than the two variants, 
showing the contribution of important sentence measurement and opinion integration. 
Key observations include:
(1) The performance of DS-Attributor(w/o IM) decreased significantly on BLEU and METEOR,
which indicates that leveraging weighted aspect vetor $v_a$ is beneficial to
remove irrelevant information and detect users' focused aspects of specific news.
(2) When opinion vector is removed, DS-Attributor(w/o OV) performs poorly on ROUGE-L and CIDEr,
which shows that mining users' opinion preference does provide prior information to understand
the sentimental tendency and thus help accurately predict comment reaction. 

\begin{table}[t]
    \setlength{\abovecaptionskip}{0.1cm}
    \centering
    \fontsize{8.5}{0}
    \begin{tabular}{p{0.95\columnwidth}}
    \toprule
    \textbf{Title:} 
    A college in Wuhan apologizes for the requisition of student dormitories: 
    Improper disposal of items will be compensated\\
    \textbf{Body:} 
    The college issued a letter of apology on February 10 in response to the requisition of students' dormitories. 
    \textcolor{blue}{The college received a notice from the city government on February 7, 
    and then requisitioned some dormitories as COVID-19 medical isolation sites by February 9.}
    The college apologized for the improper disposal of students' belongings and 
    promised to compensate students for any loss of belongings after verification 
    and disinfect the dormitories in the next semester.
    Experts: medical support. 
    In recent days, a number of university dormitories in Wuhan have been requisitioned as 
    quarantine observation points in response to the COVID-19 outbreak. 
    For students' personal belongings, many schools said they would be sealed up for special storage.\\
    \midrule
    \textbf{Seq2Seq:}
    \begin{CJK*}{UTF8}{gbsn}\fontsize{8.5}{0}我就想知道是什么时候的？\end{CJK*}
    (I just want to know when?)\\
    \textbf{Graph2Seq:} 
    \begin{CJK*}{UTF8}{gbsn}\fontsize{8.5}{0}我觉得这就是在黑，因为我觉得是个什么原因\end{CJK*}
    (I think it's slander because I think it's a reason)\\
    \textbf{Hierarchical-Attention:} 
    \begin{CJK*}{UTF8}{gbsn}\fontsize{8.5}{0}这是要被封了吗？\end{CJK*}
    (Is this going to be blocked?)\\
    \midrule
    \textbf{Comment-1:}
        \begin{CJK*}{UTF8}{gbsn}\fontsize{8.5}{0}全国学校都是一个样，都是血泪了\end{CJK*}
        Schools all over the country are the same, sad\\
    \textbf{Comment-2:}
        \begin{CJK*}{UTF8}{gbsn}\fontsize{8.5}{0}什么时候可以开学？\end{CJK*}
        When can I go to school?\\
    \textbf{Comment-3:}
        \begin{CJK*}{UTF8}{gbsn}\fontsize{8.5}{0}大逆不道！我想见宿舍！\end{CJK*}
        Outrageous! I want to see the dorm! \\
    \bottomrule   
    \end{tabular}
    \caption{
        Illustration of generated comments for one example news:
        (top) the example news; 
        (middle) generated comments by baseline methods;
        (bottom) generated comments by the proposed DS-Attributor for three different users.
        The news is originally in Chinese and translated to English and 
        the generated comments are originally in Chinese and translated to English.
    }
    \label{tab:case-study-example}
    \vspace{-2mm}
\end{table}



\begin{table}[h]
    \setlength{\abovecaptionskip}{0.1cm}
    \centering
    \resizebox{0.43\textwidth}{!}{
    \begin{tabular}{lc}
      \toprule 
      Topic No. & Topic words\\ 
      \midrule
       Aspect 1 & fan, star, hero, entertainment \\
       Aspect 4 & player, football, champion, fans, team\\
       Aspect 17 & news, society, problem, comments\\
       Aspect 20 & virus, human, earth, Black, Wuhan \\
       Aspect 38 & teacher, school, student, university \\
       Aspect 39 & world, people, politics, protest, danger\\  
      \bottomrule
    \end{tabular}
    }
    \caption{
        Example of discovered aspect topics.
    }
    \label{tab:aspect topics}
    \vspace{-2mm}
\end{table}
\begin{table}[h]
    \setlength{\abovecaptionskip}{0.1cm}
    \centering
    \resizebox{0.43\textwidth}{!}{
    \begin{tabular}{lc}
      \toprule 
      Topic No. & Topic words\\ 
      \midrule
       Opinion 1 & like, not bad, nice, pretty, delicious\\
       Opinion 2 & hope, protect, isolate, normal, alive\\  
       Opinion 3 & development, hope, try hard, solve\\ 
       Opinion 4 & no, no way, not enough, disbelief\\
      \bottomrule
    \end{tabular}
    }
    \caption{
        Example of discovered opinion topics.
    }
    \label{tab:opinion topics}
    \vspace{-2mm}
\end{table}

\subsection{Case Study}
In order to better understand how dispositional and situational attribution contribute to the comment behavior, 
as shown in Table \ref{tab:case-study-example},
we visualize the generated comments for one specific news regarding 
\emph{the requisition of a university in Wuhan as a medical isolation place.}
Compared to the baselines only leveraging modeling the situational factors, 
DS-Attributor can generate appropriate comments according to different users
considering the dispositional factors mined from their interaction history. 
It is shown that the generated comments for three example users from DS-Attributor contain diverse and more clear focuses. 
Regarding comment-3 which expresses complaint about the decision of dormintory requisition,
we examine the corresponding situational and dispositional factors. 
In particular, for situational factor, we highlight the news sentence with blue color that
users focus most on with the highest attention value $g_i$.
From situational attribution we can see that the user is concerned about dormitories and personal belongings in this news.
For dispositional factor, 
from the aspect and opinion distributions $\widehat{z}_a$, $\widehat{z}_s$, 
we find that this user has the highest preference for Aspect\#38 and Opinion\#4 . 
As shown in Table \ref{tab:aspect topics} and Table \ref{tab:opinion topics}, 
Aspect\#38 talks about school, and Opinion\#4 indicates negative sentiment.
This gives rise to the dissatisfaction in the comment and helps detect the user's actual focus in the news.

\section{Applications}
By situational and dispositional attribution, the proposed DS-Attributor can enable applications other than comment generation. In this section, 
we introduce two possible applications by employing the learned during situational and dispositional attribution.
\subsection{News Aspect-Opinion Forcasting}
In this subsection, we introduce a useful application
aggregating the predicted comments to forecast the audience's focus and opinion for future news. 
We will use an example of news to illustrate this application which describes 
\emph{Vietnam exposed a large-scale protest in the streets of the people.}
Specifically, for the given news, 
200 users are selected as test subjects to predict their focused aspect distribution $\widehat{z}_a$ and corresponding opinion distribution $\widehat{z}_s$.
For simplicity, we obtain above two topics distribution of 200 users, and analyse the topics with the highest weights respectively.

\begin{figure}[t]
    \centering
    \setlength{\abovecaptionskip}{0.1cm}
    \setlength{\belowcaptionskip}{0cm}
    \includegraphics[scale=0.32]{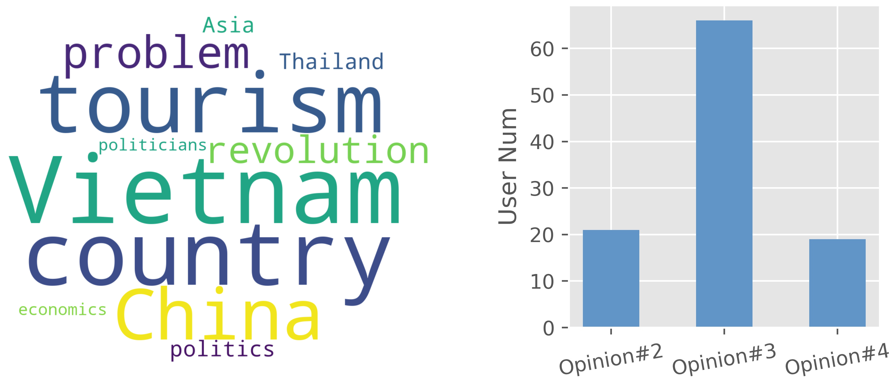}
    \caption{News aspect-opinion forecasting result.}
    \label{fig:aspect-opinion forecasting}
    \vspace{-2mm}
\end{figure}

We observe that 
most people concentrated on Aspect\#39 which talks about social topics(e.g., politics, pretest, etc.).
The keywords of generated comments on Aspect\#39 are shown in the Figure~\ref{fig:aspect-opinion forecasting}(left), 
which are closely related to the news content.
As for people's opinions on Aspect\#39,
we visualize the opinion distribution in the Figure~\ref{fig:aspect-opinion forecasting}(right).
Specifically, most people express positive emotions, such as ``protect'', ``alive'' in Opinion\#2 and
``hope'', ``solve'' in Opinion\#3,
and the others express opposition to this matter(e.g., no, disbelief, etc.).
Therefore, DS-Attributor provides the possibility to predict people's reaction before or at the early stage of news release.
By examining the users from certain community,
By examining the users from a certain community,
we can also support fine-grained aspect-opinion forecast. This will enable timely and effective public opinion management.  
\subsection{Reader-aware News Summarization}

DS-Attributor derives users' aspect preferences as by-product, 
which helps understand the subjective focus on news.
Therefore, instead of objective news summarization as most current studies
conduct by only analyzing the correlation between news sentences, 
we can exploit the derived user aspect preference to support a novel subjective news summarization. 
Specifically, we introduce subjective user factors into the traditional objective news summarization solution, by fusing
the score $g_i$ (see section \ref{sec:Situational Factor Encoder}) 
and the decoder attention $e_{i}$ (see section \ref{sec:Dynamic Decoder}) to 
update the similarity matrix $w(i,j)$ of standard TextRank~\cite{mihalcea2004textrank} as
\begin{equation}
    w(i,j) = \alpha_1 w_s(i,j) + \alpha_2 w_g(i,j) + \alpha_3 w_e(i,j),
    \label{equ:final_weight}
\end{equation}
where $w_s(i,j)$ is cosine similarity of two sentence vectors,
$w_g(i,j)$ is defined as
\begin{equation}
    w_g(i,j) = \frac{g_j w_s(i,j)} {\sum_{k} g_k w_s(i,k)},
    \label{equ:gate_weight}
\end{equation}
$w_e(i,j)$ is defined similar to $w_e(i,j)$, and $\alpha_1$, $\alpha_2$, $\alpha_3$ are coefficients.
The final sentence importance score is estimated after performing TextRank.
With ROUGE-L as the evaluation metric,
we compared three summarization strategies on 100 news:
(1) Standard TextRank: to extract top-k sentences as summary without reader factors. 
(2) Single-user: to randomly select one from 20 users' top-k results $m$ times, 
and average the ROUGE-L scores.
(3) Multi-user: to randomly select $n$ users' summary results
and choose sentences by voting each time, repeat $m$ times and average the ROUGE-L scores.

\begin{figure}[t]
    \centering 
    \setlength{\abovecaptionskip}{0.1cm}
    \includegraphics[scale=0.51]{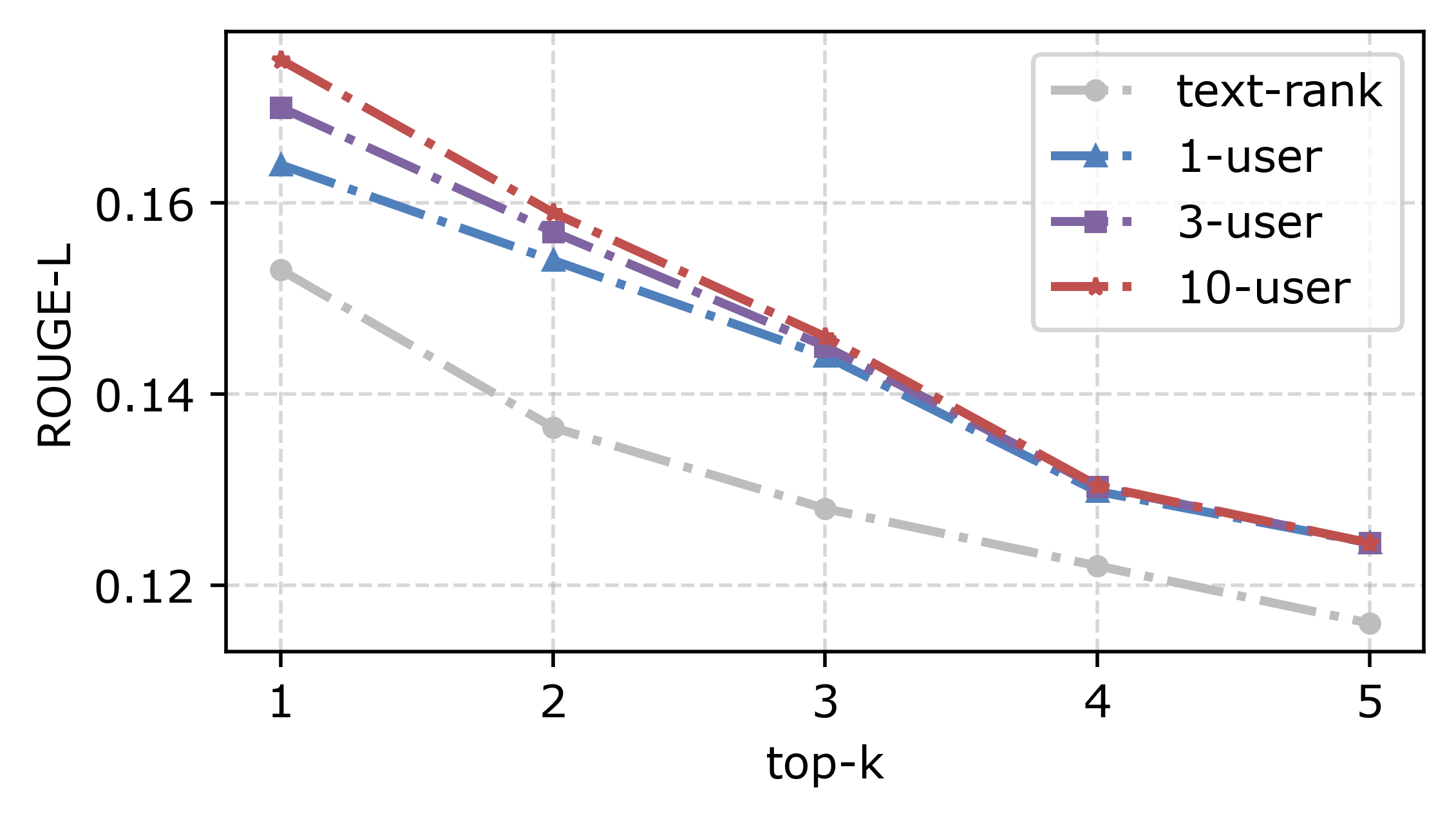}
    \caption{Reader-aware news summarization result.}
    \label{fig:summary}
    \vspace{-2mm}
\end{figure}

The evaluation results of different methods are shown in Figure \ref{fig:summary}.
From the results, we have the following conclusions:
(1) reader-aware summarization strategies show better performance than standard TextRank,
because subjective reader factor is useful to extract the news article highlight.
(2) the multi-reader strategy achieves superior performance when $k$ is small,
which shows that the common interest of multiple readers is beneficial to mining the main purpose of the news.
Users' interests disperse as $k$ increases, 
where multi-user strategy obtains close performance to single-user strategy 
but still clearly outperforms the standard TextRank-based solution. 
Note that this evaluation is conducted with news title as the ground-truth. 
In practical scenarios, by exploiting the dispositional preference for a specific individual or group of users, 
we can develop applications like customized and even personalized news summarization.


\section{Conclusion}
In this paper, we have proposed an encoder-decoder framework,
DS-Attributor, for modeling the comment generation process
by combining both situational and dispositional factors.
Following this study, we are working towards the following directions:
(1) modeling comment attribution with news event, 
e.g., associating the discovered global aspect topics to the local news event aspects; 
(2) exploring more applications by employing the derived situational and dispositional factors, 
e.g., customized news summarization, comment-driven news recommendation.

\bibliographystyle{named}
\bibliography{ijcai22}

\begin{thebibliography}{}

\bibitem[\protect\citeauthoryear{Banerjee and Lavie}{2005}]{banerjee2005meteor}
Satanjeev Banerjee and Alon Lavie.
\newblock Meteor: An automatic metric for mt evaluation with improved
  correlation with human judgments.
\newblock In {\em Proceedings of the acl workshop on intrinsic and extrinsic
  evaluation measures for machine translation and/or summarization}, pages
  65--72, 2005.

\bibitem[\protect\citeauthoryear{Boltu{\v{z}}i{\'c} and
  {\v{S}}najder}{2014}]{boltuvzic2014back}
Filip Boltu{\v{z}}i{\'c} and Jan {\v{S}}najder.
\newblock Back up your stance: Recognizing arguments in online discussions.
\newblock In {\em Proceedings of the First Workshop on Argumentation Mining},
  pages 49--58, 2014.

\bibitem[\protect\citeauthoryear{Dieng \bgroup \em et al.\egroup
  }{2020}]{dieng2020topic}
Adji~B Dieng, Francisco~JR Ruiz, and David~M Blei.
\newblock Topic modeling in embedding spaces.
\newblock {\em Transactions of the Association for Computational Linguistics},
  8:439--453, 2020.

\bibitem[\protect\citeauthoryear{Heider and Simmel}{1944}]{Heider1944AnES}
Fritz Heider and Marianne~L. Simmel.
\newblock An experimental study of apparent behavior.
\newblock {\em American Journal of Psychology}, 57:243--259, 1944.

\bibitem[\protect\citeauthoryear{Heider}{2013}]{heider2013psychology}
Fritz Heider.
\newblock {\em The psychology of interpersonal relations}.
\newblock Psychology Press, 2013.

\bibitem[\protect\citeauthoryear{Hou \bgroup \em et al.\egroup
  }{2017}]{hou2017learning}
Lei Hou, Juanzi Li, Xiao-Li Li, Jie Tang, and Xiaofei Guo.
\newblock Learning to align comments to news topics.
\newblock {\em ACM Transactions on Information Systems (TOIS)}, 36(1):1--31,
  2017.

\bibitem[\protect\citeauthoryear{Kingma and Ba}{2014}]{kingma2014adam}
Diederik~P Kingma and Jimmy Ba.
\newblock Adam: A method for stochastic optimization.
\newblock {\em arXiv preprint arXiv:1412.6980}, 2014.

\bibitem[\protect\citeauthoryear{Li \bgroup \em et al.\egroup
  }{2019}]{li2019coherent}
Wei Li, Jingjing Xu, Yancheng He, Shengli Yan, Yunfang Wu, et~al.
\newblock Coherent comment generation for chinese articles with a
  graph-to-sequence model.
\newblock {\em arXiv preprint arXiv:1906.01231}, 2019.

\bibitem[\protect\citeauthoryear{Lin}{2004}]{lin2004rouge}
Chin-Yew Lin.
\newblock Rouge: A package for automatic evaluation of summaries.
\newblock In {\em Text summarization branches out}, pages 74--81, 2004.

\bibitem[\protect\citeauthoryear{Mihalcea and
  Tarau}{2004}]{mihalcea2004textrank}
Rada Mihalcea and Paul Tarau.
\newblock Textrank: Bringing order into text.
\newblock In {\em Proceedings of the 2004 conference on empirical methods in
  natural language processing}, pages 404--411, 2004.

\bibitem[\protect\citeauthoryear{Papineni \bgroup \em et al.\egroup
  }{2002}]{papineni2002bleu}
Kishore Papineni, Salim Roukos, Todd Ward, and Wei-Jing Zhu.
\newblock Bleu: a method for automatic evaluation of machine translation.
\newblock In {\em Proceedings of the annual meeting of the Association for
  Computational Linguistics}, pages 311--318, 2002.

\bibitem[\protect\citeauthoryear{Peng \bgroup \em et al.\egroup
  }{2020}]{Peng2020KnowingWH}
Haiyun Peng, Lu~Xu, Lidong Bing, Fei Huang, Wei Lu, and Luo Si.
\newblock Knowing what, how and why: A near complete solution for aspect-based
  sentiment analysis.
\newblock In {\em AAAI}, 2020.

\bibitem[\protect\citeauthoryear{Pontiki \bgroup \em et al.\egroup
  }{2016}]{pontiki2016semeval}
Maria Pontiki, Dimitrios Galanis, Haris Papageorgiou, Ion Androutsopoulos,
  Suresh Manandhar, Mohammad Al-Smadi, Mahmoud Al-Ayyoub, Yanyan Zhao, Bing
  Qin, Orph{\'e}e De~Clercq, et~al.
\newblock Semeval-2016 task 5: Aspect based sentiment analysis.
\newblock In {\em International workshop on semantic evaluation}, pages 19--30,
  2016.

\bibitem[\protect\citeauthoryear{Qin \bgroup \em et al.\egroup
  }{2018}]{qin2018automatic}
Lianhui Qin, Lemao Liu, Victoria Bi, Yan Wang, Xiaojiang Liu, Zhiting Hu, Hai
  Zhao, and Shuming Shi.
\newblock Automatic article commenting: the task and dataset.
\newblock {\em arXiv preprint arXiv:1805.03668}, 2018.

\bibitem[\protect\citeauthoryear{Vaswani \bgroup \em et al.\egroup
  }{2017}]{vaswani2017attention}
Ashish Vaswani, Noam Shazeer, Niki Parmar, Jakob Uszkoreit, Llion Jones,
  Aidan~N Gomez, {\L}ukasz Kaiser, and Illia Polosukhin.
\newblock Attention is all you need.
\newblock In {\em Advances in neural information processing systems}, pages
  5998--6008, 2017.

\bibitem[\protect\citeauthoryear{Vedantam \bgroup \em et al.\egroup
  }{2015}]{vedantam2015cider}
Ramakrishna Vedantam, C~Lawrence~Zitnick, and Devi Parikh.
\newblock Cider: Consensus-based image description evaluation.
\newblock In {\em Proceedings of the IEEE conference on computer vision and
  pattern recognition}, pages 4566--4575, 2015.

\bibitem[\protect\citeauthoryear{Wang \bgroup \em et al.\egroup
  }{2021}]{wang2021generating}
Wei Wang, Piji Li, and Hai-Tao Zheng.
\newblock Generating diversified comments via reader-aware topic modeling and
  saliency detection.
\newblock {\em arXiv preprint arXiv:2102.06856}, 2021.

\bibitem[\protect\citeauthoryear{Yan \bgroup \em et al.\egroup
  }{2021}]{yan2021unified}
Hang Yan, Junqi Dai, Xipeng Qiu, Zheng Zhang, et~al.
\newblock A unified generative framework for aspect-based sentiment analysis.
\newblock {\em arXiv preprint arXiv:2106.04300}, 2021.

\bibitem[\protect\citeauthoryear{Yang \bgroup \em et al.\egroup
  }{2016}]{yang2016hierarchical}
Zichao Yang, Diyi Yang, Chris Dyer, Xiaodong He, Alex Smola, and Eduard Hovy.
\newblock Hierarchical attention networks for document classification.
\newblock In {\em Proceedings of the 2016 conference of the North American
  chapter of the association for computational linguistics: human language
  technologies}, pages 1480--1489, 2016.

\bibitem[\protect\citeauthoryear{Yang \bgroup \em et al.\egroup
  }{2020}]{yang2020predicting}
Fan Yang, Eduard Dragut, and Arjun Mukherjee.
\newblock Predicting personal opinion on future events with fingerprints.
\newblock In {\em Proceedings of the International Conference on Computational
  Linguistics}, pages 1802--1807, 2020.

\end{thebibliography}
\end{document}